\documentclass[lettersize,journal]{IEEEtran}
\usepackage{amsmath,amsfonts}
\usepackage{algorithmic}
\usepackage{algorithm}
\usepackage{array}
\usepackage[caption=false,font=normalsize,labelfont=sf,textfont=sf]{subfig}
\usepackage{textcomp}
\usepackage{url}
\usepackage{verbatim}
\usepackage{graphicx}
\usepackage[nocompress]{cite}
\usepackage[colorlinks=true, citecolor=blue, linkcolor=blue, urlcolor=blue]{hyperref}
\usepackage{color}
\usepackage{booktabs}
\usepackage{multirow}
\usepackage{placeins}
\begin{document}
\title{Learnable Shape Prototypes with Occlusion-Geometry-Guided Injection for Amodal Instance Segmentation}

\author{Fufan~Zhang,
        Jingxiang~Wang,
        and~Xiangjie~Ye%
\thanks{Manuscript received Month DD, 2026.}%
\thanks{Fufan Zhang and Jingxiang Wang are with the School of Mechanical Engineering and Automation, Northeastern University, Shenyang 110819, China (e-mail: zhangff1@mails.neu.edu.cn; wangjx15@mails.neu.edu.cn).}%
\thanks{Xiangjie Ye is with the School of Information Science and Engineering, Northeastern University, Shenyang 110819, China (e-mail: yexj@mails.neu.edu.cn).}%
\thanks{Corresponding author: Fufan Zhang.}}

\markboth{IEEE Transactions on Circuits and Systems for Video Technology,~Vol.~XX, No.~X, Month~2026}%
{Zhang \MakeLowercase{\textit{et al.}}: Learnable Shape Prototypes for Amodal Instance Segmentation}

\maketitle

\begin{abstract}
Amodal instance segmentation aims to predict the complete object mask including occluded regions that lack pixel-level observations and must be inferred with the aid of shape priors. Existing methods acquire shape priors through fixed-capacity encoding spaces or expensive generative models, and inject them uniformly across all spatial positions without adapting to the varying prior demand between visible and occluded regions. In this paper, we propose a gated reliability-adaptive shape prior framework, which introduces a shape prior memory module that combines learnable prototypes via cross-attention to produce instance-adaptive shape priors through weighted prototype combination rather than generation. A spatial adaptive reliability gate then employs the signed distance field of the visible mask to modulate injection intensity at each position according to its occlusion depth, preserving reliable features in visible regions while directing shape compensation toward occluded areas. Experiments on two mainstream amodal instance segmentation benchmarks demonstrate that the proposed method outperforms existing approaches under multiple evaluation settings, improving the mean intersection-over-union over occluded regions by over 11 percentage points on one of the two benchmarks under the standard setting, while using approximately one-third of the total parameters. Linear probing analysis further reveals that the visible-mask cross-attention module implicitly encodes occlusion geometry into visual token representations, explaining the effectiveness of the proposed module decomposition.
\end{abstract}



\section{Introduction}

\begin{figure}[t]
\centering
\includegraphics[width=\linewidth]{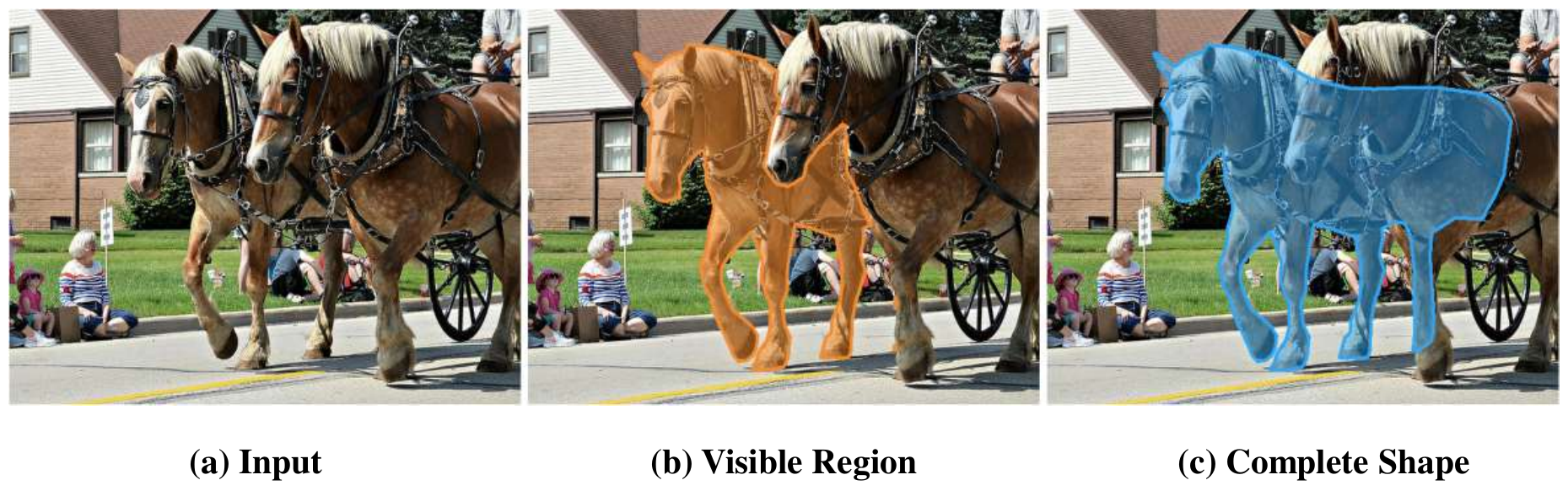}
\caption{Illustration of the amodal instance segmentation task. Given an RGB input image (left), the visible mask (middle) captures only the observable portion of each object, while the amodal mask (right) additionally recovers the occluded regions that lack pixel-level evidence in the input.}
\label{fig:task}
\end{figure}

\IEEEPARstart{A}{modal} instance segmentation aims to predict the complete mask of each object in an image, covering both the visible portion and the part occluded by other objects~\cite{li2016amodal, zhu2017semantic, qi2019amodal}. As illustrated in Fig.~\ref{fig:task}, given an input image and the visible region mask of an object, the task requires the model to infer the full object contour including the occluded portion. This capability has broad practical value in visual understanding tasks such as the perception of occluded vehicles in autonomous driving~\cite{qi2019amodal, ke2023bcnet, follmann2019learning}, the localization and grasping of partially occluded objects in robotic manipulation~\cite{mohan2022amodal, tai2025segment}, and the parsing of object-level layering in scene editing and content generation~\cite{zhan2020deocclusion, ao2025openworld, zhu2025entityerasure}. The task has received sustained attention in the computer vision community~\cite{xiao2021amodal, aisformer2022, oaformer2023, sun2022bayesian}. Unlike conventional instance segmentation~\cite{he2017maskrcnn, cheng2021maskformer, cheng2022mask2former, jain2023oneformer, kirillov2023sam}, which only needs to delineate visible pixels, amodal segmentation faces the fundamental challenge that occluded regions lack any direct pixel-level observation in the image, making it impossible for the model to infer their shape from visual input alone. To recover the contours of these invisible regions, the model must incorporate prior knowledge about the typical shapes of objects, commonly referred to as shape priors~\cite{xiao2021amodal, gao2023c2f, gin2024, aisformer2022, blade2024}. The manner in which shape priors are acquired and utilized largely determines the performance of amodal segmentation methods.

Existing research on shape prior acquisition has proceeded along two directions. One line of methods learns and stores typical object shapes from training data, and at inference time uses the stored shapes as references to assist in predicting the invisible regions~\cite{xiao2021amodal, gao2023c2f, gin2024, aisformer2022, blade2024}. This approach has evolved from early fixed-dictionary shape retrieval~\cite{xiao2021amodal}, through vector-quantized latent space modeling~\cite{gao2023c2f, esser2021taming}, to transformation-invariant shape encoding~\cite{gin2024}, progressively improving the expressiveness of the shape prior. However, the expressiveness of these methods is constrained by the discreteness of the encoding space, limiting coverage of the full shape variation within each category~\cite{gin2024, blade2024}. Another line of methods draws on large-scale pretrained generative models~\cite{rombach2022ldm} to reformulate shape prediction of invisible regions as a conditional image generation problem~\cite{ozguroglu2024pix2gestalt, liu2025samba, chen2025diffusion, tran2024diffsp}. Benefiting from the visual distribution knowledge learned from large-scale data, generative approaches exhibit strong shape imagination capabilities and demonstrate clear advantages in zero-shot scenarios~\cite{ozguroglu2024pix2gestalt, liu2025samba}. Nevertheless, the computational cost of these methods is considerably higher than that of retrieval-based alternatives, and they typically rely on ground-truth-level input conditions at test time~\cite{ozguroglu2024pix2gestalt, liu2025samba}, which limits their applicability in practical deployment. How to achieve sufficiently flexible shape prior representations while keeping computational costs under control remains an open problem in this field.

Beyond the acquisition of shape priors, the strategy by which shape priors are injected into visual features also has an important influence on final performance. We observe that both categories of methods described above share a common limitation in this regard: they apply shape priors at uniform intensity across all spatial locations~\cite{xiao2021amodal, gao2023c2f, gin2024, ozguroglu2024pix2gestalt, liu2025samba}. However, the actual demand for shape prior assistance varies significantly across different spatial positions. In the visible regions of an object, current visual foundation models~\cite{dosovitskiy2021vit, caron2021dino, he2022mae, oquab2024dinov2, liu2021swin} are already capable of extracting high-quality feature representations from real pixels~\cite{oaformer2023}, and superimposing shape priors at these locations may interfere with the already reliable features. In contrast, at positions far from the visible boundary and deep within the occluded area, the image provides no usable visual cues, and shape priors become the primary basis for prediction~\cite{xiao2021amodal, gin2024}. Near the occlusion boundary, visible contours can still offer local continuation information, consistent with the relatability principle in visual completion~\cite{ke2023bcnet, zhang2019sln, kellman1991relatability}, and the required level of prior assistance lies between the two extremes. In other words, the difficulty of shape completion is spatially non-uniform~\cite{kellman1991relatability}, yet existing methods do not reflect this variation in their injection strategy.

To address these two problems, we propose GRASP (Gated Reliability-Adaptive Shape Prior), a shape prior framework for amodal instance segmentation. For shape prior acquisition, GRASP introduces a Shape Prior Memory (SPM) module that combines multiple learnable prototypes via cross-attention to produce instance-specific shape priors. Unlike discrete retrieval methods that select one or a few entries from a fixed codebook~\cite{xiao2021amodal, gao2023c2f, gin2024}, SPM performs soft combination over the entire prototype bank, yielding a continuous representation space more flexible than discrete codebook selection. This design preserves the computational efficiency of codebook-based approaches while avoiding the capacity bottleneck of discrete codebooks. For shape prior utilization, GRASP develops a Spatial Adaptive Reliability Gate (Spatial ARG) that employs the signed distance field (SDF)~\cite{zhang2019sln} of the visible mask to assign position-wise prior injection intensities. At the decoding stage, a Hybrid Fused Dual-Head Decoder (HFDHD) decomposes amodal mask prediction into visible and occluded branches~\cite{qi2019amodal, ke2023bcnet}. The entire framework uses a frozen DINOv2~\cite{oquab2024dinov2} as the visual encoder, with only approximately 26M trainable parameters.

The main contributions of this paper can be summarized as follows.

\begin{enumerate}
\item We propose Shape Prior Memory (SPM), a lightweight cross-attention-based module that combines shape completion information from a learnable prototype bank, replacing costly generative shape priors with substantially fewer trainable parameters.

\item We introduce Spatial Adaptive Reliability Gate (Spatial ARG), which employs the signed distance field of the visible mask to adaptively allocate shape prior corrections, directing compensation toward occluded regions while preserving visible-region features. Ablation studies confirm its significant contribution to occluded-region segmentation accuracy.

\item Extensive experiments on two benchmarks demonstrate that GRASP achieves competitive accuracy with substantially fewer trainable parameters and faster inference than codebook-based alternatives. We additionally reveal through linear probing that VM Cross-Attention implicitly encodes occlusion geometry, indicating that explicit geometric injection at the prototype query is unnecessary when visible-mask cross-attention is already present in the pipeline.
\end{enumerate}

The remainder of this paper is organized as follows. Section~II reviews related work on amodal instance segmentation. Section~III presents the proposed GRASP framework in detail. Section~IV reports experimental results and analysis. Section~V concludes the paper.


\section{Related Work}

\subsection{Architectural Approaches for Amodal Segmentation}

The methodology of amodal instance segmentation has undergone continuous evolution centered on occlusion-aware network design. Early methods followed the two-stage detect-then-segment paradigm of standard instance segmentation~\cite{he2017maskrcnn, cheng2021maskformer, cheng2022mask2former}, predicting complete object masks within detection bounding boxes~\cite{li2016amodal, follmann2019learning}, while concurrent work explored adversarial generation to recover occluded content~\cite{ehsani2018segan}. Since bounding boxes frequently contain multiple overlapping instances and a target mask may extend beyond its detected box~\cite{ke2021bcnet, ke2023bcnet, zhang2022beyondbbox}, subsequent work introduced occlusion-aware designs including bilayer feature decomposition into occluder and occludee representations~\cite{ke2021bcnet, ke2023bcnet}, and auxiliary branches that jointly predict occlusion status~\cite{qi2019amodal, oaformer2023}, visible region masks~\cite{follmann2019learning, zhang2024opnet}, occlusion boundaries~\cite{zhang2019sln, zhang2024opnet}, and signed-distance-field-based encodings~\cite{zhang2019sln, nguyen2021weakly}, enhancing occlusion relationship modeling through multi-task learning~\cite{chen2023priorguidedexpansion, back2022unseen}.

More recently, Transformer architectures~\cite{vaswani2017attention} have been introduced to this task, enabling end-to-end modeling of the relationships among visible, occluded, and amodal masks via attention mechanisms~\cite{aisformer2022, tran2024shapeformer}. In the broader instance segmentation community, unified frameworks have been developed to handle detection and segmentation jointly~\cite{li2023maskdino, jain2023oneformer}, with parallel research exploring contour-based representations that iteratively refine a polygon toward object boundaries as a compact alternative to dense pixel-wise masks~\cite{feng2024polysnake}, and these architectural advances inform the design space for amodal methods as well. The research scope has also expanded to amodal panoptic segmentation~\cite{mohan2022amodal}, dense intra-class occlusion~\cite{ao2024intraclass}, temporal amodal reasoning~\cite{chen2025diffusion}, weakly supervised settings~\cite{blade2024}, and occlusion-aware extensions of visual foundation models~\cite{tai2025segment, kirillov2023sam, liu2025samba, oquab2024dinov2}. The above efforts primarily advance the segmentation architecture itself, whereas this paper focuses on shape prior acquisition and injection, two problems complementary to architectural design.

\subsection{Shape Prior for Occluded Region Completion}

Occluded regions lack direct pixel-level observations, and models must rely on shape priors to infer the contours of invisible parts~\cite{xiao2021amodal, gao2023c2f, gin2024}. The general principle of using a reference signal to guide segmentation has also proven effective in adjacent dense-prediction tasks, such as anomaly detection where normal-image features serve as guidance for identifying deviating regions~\cite{xing2024nigsf}. The representational form of shape priors has evolved from discrete templates through latent-space encoding to continuous generative models in existing work~\cite{xiao2021amodal, gao2023c2f, gin2024, ozguroglu2024pix2gestalt, liu2025samba}.

Early work constructed category-level shape dictionaries, encoding the typical contours of each object class as a set of discrete prototype vectors and matching shape references to target instances via nearest-neighbor retrieval at inference time~\cite{xiao2021amodal}. Beyond static dictionaries, memory-network paradigms in dense video prediction have demonstrated the value of cross-frame feature retrieval for handling occlusion and appearance variation~\cite{chen2024memnet}. The introduction of vector quantization techniques~\cite{oord2017vqvae} mapped object shapes into a discrete latent codebook space~\cite{gao2023c2f, esser2021taming}, enabling more flexible shape reconstruction through the combination of multiple codebook entries~\cite{gao2023c2f}. Follow-up work further incorporated transformation-invariance constraints to improve encoding robustness~\cite{gin2024}, while parallel research in few-shot segmentation has shown that learnable prototypes combined with query-guided attention can adaptively combine reference features for novel instances~\cite{hu2024prototypes}, or exploited category information to construct class-specific codebooks for higher retrieval precision~\cite{tran2024shapeformer}.

Recent work has turned to pretrained diffusion models~\cite{rombach2022ldm} for shape prior acquisition, formulating occluded region completion as a conditional generation process~\cite{ozguroglu2024pix2gestalt, tran2024diffsp}. Within this paradigm, considerable technical diversity exists. One branch operates in the RGB image space, fine-tuning a pretrained Stable Diffusion model with dual-stream conditioning from CLIP semantic encodings and VAE visual encodings of the occluded image, first synthesizing an image of the complete object and then extracting the amodal mask from it~\cite{ozguroglu2024pix2gestalt}. Another branch works directly in a more compact mask space, bypassing the computational burden of high-resolution image generation~\cite{tran2024diffsp}. In video settings, diffusion models have been further extended into a temporal conditional generation framework, taking modal mask sequences and pseudo-depth maps as input and propagating shape information across frames via spatiotemporal priors of video generative models to maintain temporal consistency~\cite{chen2025diffusion}. Large-scale pretrained segmentation foundation models have also been adapted to the amodal task, enhancing occluded region reasoning through joint modeling of visible and amodal segmentation~\cite{liu2025samba}.

While the representational capacity of shape priors has improved steadily through these advances, each category of methods faces its own limitations. Discrete dictionaries and codebooks have a fixed capacity~\cite{xiao2021amodal, gin2024}, and their expressiveness remains constrained by the discreteness of the encoding space, with noticeable performance degradation when shape diversity exceeds codebook coverage~\cite{gin2024, blade2024}. Generative methods typically require multi-step denoising iterations or large-scale model backbones at inference time, incurring a computational overhead that is orders of magnitude higher than retrieval-based approaches~\cite{ozguroglu2024pix2gestalt, liu2025samba}. The SPM module proposed in this paper adopts learnable prototypes combined with cross-attention to realize lightweight shape prior construction, achieving flexible shape expressiveness at the low computational cost of codebook-based methods.

\subsection{Adaptive Feature Modulation and Gating}

Adapting the injection intensity of shape priors to the spatial characteristics of each position requires position-wise weight allocation, for which attention and gating mechanisms provide a natural technical foundation~\cite{hu2018senet, woo2018cbam, perez2018film, xia2022dat, yang2022focalmod}. In the channel dimension, learning an adaptive weight for each feature channel has become a standard practice~\cite{hu2018senet}, with subsequent improvements reducing parameter overhead~\cite{wang2020ecanet} and incorporating positional encoding~\cite{hou2021coordattn}. In the spatial dimension, generating per-location weight maps to guide the network toward task-relevant regions has likewise proven effective~\cite{woo2018cbam, xia2022dat, cheng2024rcsanet}, and the two dimensions of attention can be applied in sequence to progressively refine features along both channel and spatial axes~\cite{woo2018cbam, hou2021coordattn}. Beyond additive or multiplicative attention, focal modulation provides an alternative paradigm that aggregates context through learnable gating around each query position~\cite{yang2022focalmod}. In the context of conditional feature modulation, per-channel affine transformations driven by external signals have been proposed for injecting task-relevant conditioning information~\cite{perez2018film}. In the image inpainting domain, similar gating strategies have been employed to enable the network to distinguish valid pixels from missing regions, suppressing responses to invalid features during the restoration process~\cite{yu2019gatedconv}.

These mechanisms have been extensively validated in image classification~\cite{hu2018senet}, object detection~\cite{woo2018cbam, xia2022dat}, and dense prediction tasks~\cite{hou2021coordattn, wang2020ecanet, yang2022focalmod}, yet they have not been applied to modulate the injection intensity of shape priors in amodal segmentation. In this setting, each spatial position's reliance on shape priors is determined by its occlusion depth, which can be quantitatively characterized through the signed distance field (SDF) of the visible mask~\cite{zhang2019sln, park2019deepsdf}. However, existing attention methods learn weight assignments in a purely data-driven framework~\cite{hu2018senet, woo2018cbam}, without incorporating occlusion geometry as an explicitly available structural signal into the gating process. The Spatial ARG proposed in this paper addresses this gap by incorporating SDF-encoded occlusion geometry directly into the gating signal, enabling position-wise control over shape prior injection.


\section{Proposed Method}

\subsection{Overview}

Given an input RGB image $I$ and the visible mask $V$ of an object instance, the goal of amodal instance segmentation is to predict the complete object mask $A$, which covers both the visible and the occluded portions. The visible mask may come from ground-truth annotations (Oracle setting) or from the output of an upstream instance segmentation model (Standard setting). The proposed GRASP framework addresses two core questions: how to acquire and inject shape prior knowledge efficiently, and how to adaptively regulate the injection intensity according to occlusion geometry.

\begin{figure*}[!t]
\centering
\includegraphics[width=\textwidth]{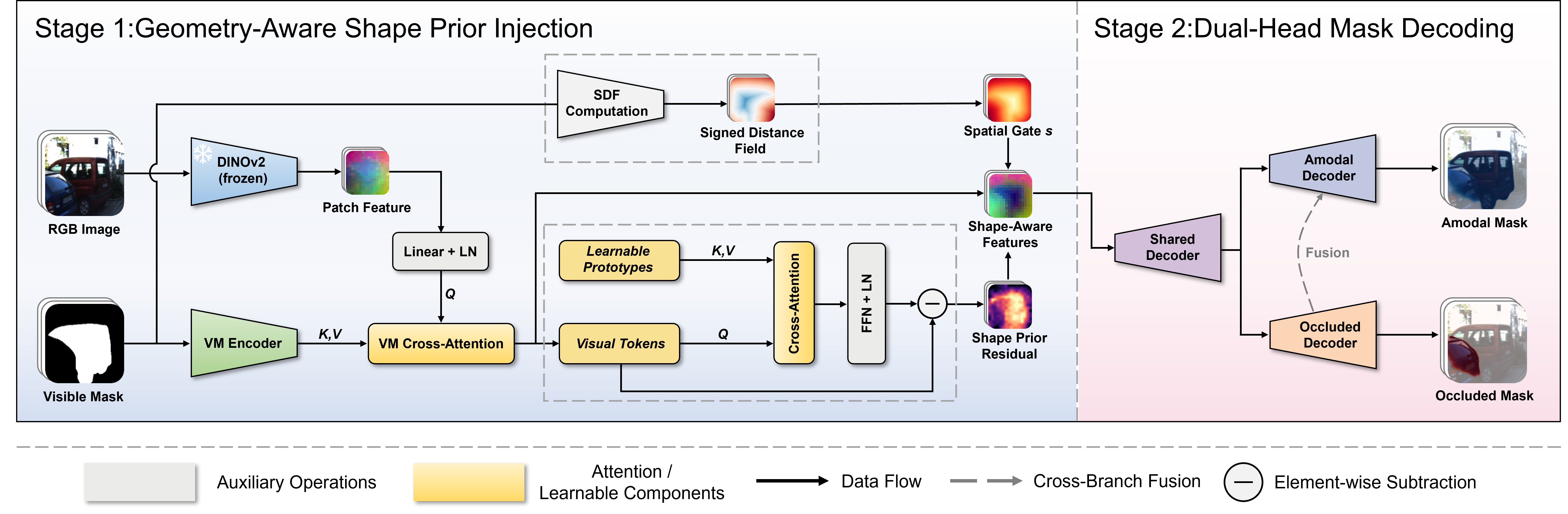}
\caption{Overall architecture of the proposed GRASP framework. The pipeline takes an RGB image and its visible mask as input and processes them through five stages: frozen DINOv2 feature extraction, VM Encoder fusion, Shape Prior Memory combination, Spatial Adaptive Reliability Gate modulation, and Hybrid Fused Dual-Head Decoder prediction. Shape prior injection is spatially modulated by the signed distance field of the visible mask.}
\label{fig:architecture}
\end{figure*}

The overall architecture of GRASP is illustrated in Fig.~\ref{fig:architecture}. The pipeline consists of five components. A frozen DINOv2-ViT-B/14~\cite{oquab2024dinov2} serves as the visual feature extractor, and its multi-layer patch tokens are concatenated and linearly projected into a unified feature representation. The VM Encoder integrates the visible mask information into the image features through residual cross-attention, enabling the model to perceive the boundary between visible and occluded regions. The Shape Prior Memory (SPM) module maintains a set of learnable shape prototypes and constructs instance-specific shape priors via cross-attention. The Spatial Adaptive Reliability Gate (Spatial ARG) transforms the shape prior injection from a spatially uniform operation into a spatially adaptive one, using signed distance field (SDF) geometry to control injection intensity at each spatial position. The Hybrid Fused Dual-Head Decoder (HFDHD) decodes the modulated features into occluded and amodal mask predictions, maintaining geometric consistency between the two outputs through a unidirectional occluded-to-amodal information flow.

\subsection{Feature Extraction and VM Encoding}

The quality of visual features directly determines the upper bound of occlusion completion. Self-supervised pretrained vision models~\cite{caron2021dino, he2022mae, oquab2024dinov2}, built upon the Vision Transformer architecture~\cite{dosovitskiy2021vit}, have demonstrated strong representation capabilities across a variety of dense prediction tasks~\cite{chen2023vitadapter, yang2024depthanything}. GRASP adopts a frozen DINOv2-ViT-B/14~\cite{oquab2024dinov2, darcet2024registers} as the backbone and does not update its parameters throughout training. Unlike methods that only use the output of the final layer, we extract patch tokens from the last four Transformer layers, concatenate them along the channel dimension, and map them to a unified dimension $D$ through a linear projection layer. This multi-layer fusion strategy allows the resulting features to simultaneously capture fine-grained spatial information from intermediate layers and high-level semantic information from deeper layers, both of which are beneficial for occluded region completion.

Effectively incorporating visible mask information into the visual features is a key step in amodal segmentation. We design a VM Encoder to accomplish this task. The VM Encoder first encodes the binary visible mask $V$ into a feature representation $F_v$ at the same spatial resolution as the patch tokens through a lightweight convolutional network. It then integrates $F_v$ into the image features $F$ via a residual cross-attention mechanism. Specifically, the image features $F$ serve as the query while the visible mask features $F_v$ serve as the key and value, and the fusion process can be expressed as
\begin{equation}
\label{eq:vm_encoder}
F' = F + \gamma \cdot \mathrm{CrossAttn}(Q{=}F,\; K{=}F_v,\; V{=}F_v),
\end{equation}
where $\gamma$ is a learnable scalar parameter initialized to zero. This zero-initialization design ensures that the cross-attention output does not disturb the high-quality visual representations already learned by DINOv2 at the beginning of training. As training progresses, $\gamma$ gradually increases and the visible mask information is smoothly introduced into the features.

In practical applications, the input visible mask may contain noise and errors. To strengthen the robustness of the model against imperfect inputs, we adopt a mixed VM noise injection strategy during training. For each training sample, we use the ground-truth visible mask with probability 0.5, and with probability 0.5 we apply a randomly perturbed version whose boundary has been randomly shifted and whose region has been randomly expanded or contracted, simulating the typical error patterns of upstream segmentation models. Similar augmentation strategies have been adopted in prior amodal segmentation work~\cite{gao2023c2f}. This approach exposes the model to both clean and noisy inputs during training, thereby maintaining tolerance to upstream errors under the Standard setting.

\subsection{Shape Prior Memory}

Occluded pixels are completely invisible in the input image, and the model therefore requires some form of prior knowledge to infer the shape of the occluded regions. Existing methods typically rely on generative priors, such as VQ-VAE codebook-based shape reconstruction~\cite{gao2023c2f,ao2024intraclass} or diffusion model-based shape synthesis~\cite{ozguroglu2024pix2gestalt}. While effective, these approaches require an independent training pipeline for prior acquisition and incur considerable computational overhead. GRASP takes a different approach by combining a set of learnable shape prototypes with a cross-attention combination mechanism to obtain shape priors. Unlike VQ-VAE codebook methods~\cite{gao2023c2f,ao2024intraclass} that require a separately trained encoder-decoder for codebook construction, the prototypes in SPM are jointly optimized with the rest of the network, eliminating the need for an additional pretraining stage and simplifying the overall training pipeline.

The module maintains a memory bank of $N_p = 128$ learnable prototypes $P \in \mathbb{R}^{N_p \times D}$, where each prototype is a $D$-dimensional vector that is updated together with the other network parameters through backpropagation during training. In the forward pass, the image features $F'$ fused by the VM Encoder serve as the query, while the prototype set $P$ simultaneously serves as the key and value, and shape prior construction is performed through standard multi-head cross-attention:
\begin{equation}
\label{eq:spm}
H = \mathrm{CrossAttn}(Q{=}F',\; K{=}P,\; V{=}P),
\end{equation}
where $H \in \mathbb{R}^{L \times D}$ denotes the combined shape prior features and $L$ is the number of patch tokens. The attention weights reflect the degree of focus that each spatial position places on different prototypes, and through learning the model can cause different prototypes to specialize in different shape patterns. It is worth noting that this mechanism performs soft combination over all prototypes rather than discrete selection of a single entry, as in VQ-VAE codebook methods~\cite{gao2023c2f, oord2017vqvae}. The resulting shape prior for each token is a weighted mixture of all 128 prototypes, producing a continuous representation space more flexible than discrete codebook selection.

To allow shape prior information to supplement the existing features incrementally rather than overwrite them, we adopt residual-style fusion. We define the shape prior residual $\Delta = H - F'$, which represents the complementary information that the combined shape prior provides relative to the current features. This residual is injected into the features in a spatially adaptive manner by the subsequent Spatial ARG, rather than added back in its entirety. This separation makes the gate control more precise: it regulates the magnitude of prior injection rather than the presence or absence of the prior itself.

From the perspective of computational efficiency, the parameter count of SPM is only $N_p \times D$ floating-point numbers, and the computational complexity of the cross-attention is $O(L \times N_p \times D)$. Since $N_p = 128$ is much smaller than the typical number of patch tokens $L$, the additional overhead introduced by SPM is negligible. Compared with independently trained VQ-VAE encoders~\cite{gao2023c2f,ao2024intraclass} or iterative sampling in diffusion models~\cite{ozguroglu2024pix2gestalt}, this approach offers clear advantages in both parameter count and inference speed.

\subsection{Spatial Adaptive Reliability Gate}

Although shape priors provide important information for completing occluded regions, not all spatial positions require these priors equally. For pixels in visible regions, the visual features extracted by DINOv2 already contain sufficient appearance and semantic information, and injecting shape priors at these locations may introduce interference. For pixels in occluded regions, the lack of direct visual evidence makes shape priors the primary basis for inferring their shape. Existing methods typically add shape priors at uniform intensity across all spatial positions~\cite{xiao2021amodal,gin2024}, overlooking the fundamental difference in prior demand between visible and occluded regions.

To address this spatially non-uniform demand, we propose the Spatial Adaptive Reliability Gate (Spatial ARG), which uses the signed distance field (SDF)~\cite{park2019deepsdf} of the visible mask to generate adaptive per-position injection weights. The SDF is defined on the visible mask and computes, for each pixel position, the signed distance to the mask boundary: positive outside the mask (occluded region), negative inside (visible region), and zero on the boundary. Formally, for a position $p$, its SDF value is defined as
\begin{equation}
\label{eq:sdf}
\mathrm{SDF}(p) =
\begin{cases}
+d(p, \partial V), & \text{if } p \notin V, \\[2pt]
-d(p, \partial V), & \text{if } p \in V,
\end{cases}
\end{equation}
where $d(p, \partial V)$ is the Euclidean distance from position $p$ to the visible mask boundary $\partial V$. To make SDF values comparable across objects of different scales, we normalize the raw SDF by the image diagonal length, yielding normalized values $\bar{d}_i \in [-1, 1]$ for each patch token position $i$. These normalized values are then mapped to spatial gate weights through a parametric sigmoid function:
\begin{equation}
\label{eq:spatial}
s_i = \sigma(\alpha \cdot \bar{d}_i + \beta),
\end{equation}
where $\alpha$ controls the sharpness of the transition between visible and occluded regions, $\beta$ controls the overall gate bias, and both are learnable scalar parameters. This minimal parameterization (only two scalars) ensures that the spatial gate is driven entirely by occlusion geometry rather than learned in a purely data-driven manner. On the KINS dataset, the model learns $\alpha = 2.68$ and $\beta = 0.26$, yielding gate values of approximately 0.08 in visible interiors, 0.56 near occlusion boundaries, and 0.95 in deeply occluded regions, indicating that two scalar parameters are sufficient to produce a spatially meaningful gate distribution aligned with occlusion depth.

\begin{figure*}[!t]
\centering
\includegraphics[width=\textwidth]{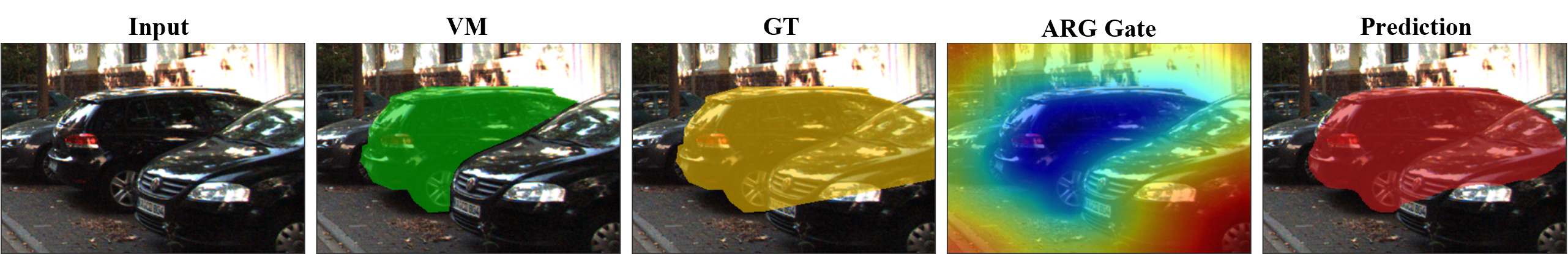}
\caption{Visualization of the Adaptive Reliability Gate. From left to right: input image, ground-truth visible mask, ground-truth amodal mask, ARG gate activation $s = \sigma(\alpha\bar{d} + \beta)$ overlaid on the input (warmer = higher), and GRASP prediction. The gate concentrates near visible--occluded boundaries, where shape prior injection from SPM is most beneficial.}
\label{fig:gate_vis}
\end{figure*}

The shape prior residual is then modulated by the spatial gate before injection:
\begin{equation}
\label{eq:gate}
F_{\mathrm{out}} = F' + s \odot \Delta,
\end{equation}
where $\odot$ denotes element-wise multiplication and $\Delta = H - F'$ is the shape prior residual from SPM. Through SDF-driven gating, shape priors are fully injected in deeply occluded regions, naturally suppressed in visible regions, and smoothly interpolated near occlusion boundaries. Although the gate also takes high values in background regions far from the object, the corresponding shape prior residual $\Delta$ vanishes there, so the effective injection $s \odot \Delta$ remains localized to object boundaries (Fig.~\ref{fig:gate_vis}).

\subsection{Hybrid Fused Dual-Head Decoder}

The modulated features $F_{\mathrm{out}}$ are decoded into final mask predictions through the Hybrid Fused Dual-Head Decoder (HFDHD). In amodal segmentation, the model must simultaneously output an amodal mask and an occluded mask, and these two masks satisfy a deterministic geometric relationship: the amodal mask equals the union of the visible mask and the occluded mask~\cite{qi2019amodal, ke2021bcnet}. HFDHD is designed to exploit this constraint through a three-stage architecture.

In the first stage, $F_{\mathrm{out}}$ passes through shared upsampling and convolutional layers that progressively restore spatial resolution while maintaining common features useful for both tasks. In the second stage, the feature stream splits into two independent branches (occluded and amodal), each containing independent convolutional layers to learn task-specific representations. In the third stage, the intermediate features of the occluded branch are concatenated with those of the amodal branch and passed through a convolutional fusion layer, forming a unidirectional occluded-to-amodal information flow. Formally, the amodal branch output is
\begin{equation}
\label{eq:amodal_head}
\hat{A} = \mathrm{Head}_a\!\bigl(\mathrm{Conv}_{\mathrm{fuse}}([F_a;\, F_o])\bigr),
\end{equation}
where $F_o$ and $F_a$ are the occluded and amodal branch features, $[{\cdot}\,;\,{\cdot}]$ denotes channel-wise concatenation, and $\mathrm{Head}_a$ is the amodal prediction head. The occluded branch output $\hat{O} = \mathrm{Head}_o(F_o)$ is produced independently without receiving feedback from the amodal branch. This asymmetric design keeps occluded prediction independent of visible-region information while allowing amodal prediction to benefit from the estimated occlusion map.

\subsection{Loss Function}

The training loss of GRASP consists of two terms. For the amodal mask prediction, we use a weighted combination of binary cross-entropy loss (BCE) and Dice loss~\cite{milletari2016vnet}:
\begin{equation}
\label{eq:loss_amodal}
\mathcal{L}_{\mathrm{amodal}} = \mathrm{BCE}(\hat{A}, A_{\mathrm{gt}}) + \mathrm{Dice}(\hat{A}, A_{\mathrm{gt}}).
\end{equation}
For the occluded mask prediction, the ground-truth occluded mask is derived as $O_{\mathrm{gt}} = A_{\mathrm{gt}} \setminus V_{\mathrm{gt}}$, i.e., the amodal region excluding the visible region. We adopt the same loss form but multiply it by a weight factor of 1.5 to address the class imbalance between occluded and visible pixels:
\begin{equation}
\label{eq:loss_occ}
\mathcal{L}_{\mathrm{occluded}} = 1.5 \times \bigl(\mathrm{BCE}(\hat{O}, O_{\mathrm{gt}}) + \mathrm{Dice}(\hat{O}, O_{\mathrm{gt}})\bigr).
\end{equation}
The elevated weight reflects the fact that occluded regions typically account for only a small fraction of the image. Without this emphasis, the model tends to predict everything as visible, which yields a low overall loss but provides no benefit for amodal completion. The total loss function is
\begin{equation}
\label{eq:loss_total}
\mathcal{L}_{\mathrm{total}} = \mathcal{L}_{\mathrm{amodal}} + \mathcal{L}_{\mathrm{occluded}}.
\end{equation}

All trainable modules (VM Encoder, VM Cross-Attention, SPM, Spatial ARG, and HFDHD) are randomly initialized and jointly optimized using the AdamW optimizer~\cite{loshchilov2019adamw}.

\subsection{Computational Complexity}

A key design goal of GRASP is to achieve flexible shape prior representation without the computational overhead of generative models. Table~\ref{tab:complexity} breaks down the parameter budget across modules.

\begin{table}[!t]
\centering
\caption{Per-module parameter breakdown of GRASP.}
\label{tab:complexity}
\renewcommand{\arraystretch}{1.15}
\begin{tabular}{l r l}
\toprule
\textbf{Module} & \textbf{Params} & \textbf{Note} \\
\midrule
DINOv2-ViT-B/14 & 86M & Frozen \\
Linear projection & 2.4M & 3072d$\to$768d \\
VM Encoder + Cross-Attn & 7.1M & Conv + single-layer CA \\
SPM (prototypes + CA) & 1.2M & $128 \times 768$ + CA layer \\
Spatial ARG & 2 & $\alpha$, $\beta$ only \\
HFDHD decoder & 15.3M & Shared + split + fusion \\
\midrule
\textbf{Total trainable} & \textbf{26M} & \\
\textbf{Total (incl.\ frozen)} & \textbf{112M} & \\
\bottomrule
\end{tabular}
\end{table}

The frozen DINOv2 backbone accounts for 86M parameters that are shared across all downstream tasks and do not participate in gradient computation. Among the 26M trainable parameters, the HFDHD decoder constitutes the largest portion (15.3M) due to its progressive upsampling structure, while SPM adds only 1.2M parameters for shape prior construction. The Spatial ARG module contains just two scalar parameters, making the geometry-driven gating mechanism essentially parameter-free. In comparison, C2F-Seg~\cite{gao2023c2f} requires 325M total parameters including a separately trained VQ-GAN encoder-decoder for codebook construction~\cite{esser2021taming} and multiple rounds of iterative MaskGIT decoding at inference time. The computational complexity of SPM cross-attention is $O(L \times N_p \times D)$, where $L{=}256$ (patch tokens), $N_p{=}128$ (prototypes), and $D{=}768$, amounting to a single matrix multiplication that completes in under 1\,ms on modern GPUs.

\subsection{Two-Pass Iterative Inference}

Under the Standard setting, the visible mask predicted by the upstream model inevitably contains errors that propagate through the pipeline. To mitigate this, GRASP employs a two-pass iterative inference strategy. In the first pass, the upstream-predicted visible mask $V_{\mathrm{pred}}$ is used as input, producing an initial amodal prediction $\hat{A}_1$ and occluded prediction $\hat{O}_1$. From these, a refined visible mask is derived as $V_{\mathrm{ref}} = \hat{A}_1 \setminus \hat{O}_1$, which typically has higher boundary accuracy than $V_{\mathrm{pred}}$ because the amodal prediction extends the object contour beyond the visible region. In the second pass, $V_{\mathrm{ref}}$ replaces $V_{\mathrm{pred}}$ as the model input, and the resulting predictions $\hat{A}_2$ and $\hat{O}_2$ serve as the final output. This strategy is analogous to the iterative MaskGIT decoding used in C2F-Seg~\cite{gao2023c2f}, but requires only two forward passes rather than multiple decoding rounds. We do not apply further iterations beyond the second pass, as Oracle-setting experiments with self-refinement show diminishing returns (full mIoU decreases by 0.62 when a third pass is applied), indicating that two passes are sufficient to capture the available information.



\section{Experiments}

\subsection{Experimental Setup}

We evaluate the proposed GRASP on two mainstream amodal instance segmentation benchmarks. KINS~\cite{qi2019amodal}, derived from the KITTI driving benchmark~\cite{geiger2012kitti}, focuses on autonomous driving scenarios, containing approximately 92K test instances across seven traffic-scene object categories dominated by vehicles, with relatively regular occlusion patterns. COCOA~\cite{zhu2017semantic}, derived from the MS-COCO dataset~\cite{lin2014coco}, covers general indoor and outdoor scenes with approximately 3.8K test instances spanning 80 object categories and more complex occlusion configurations.

Following the standard evaluation protocol in this field~\cite{qi2019amodal, ke2023bcnet, gao2023c2f, gin2024}, we report two metrics: full mIoU, which measures the intersection-over-union between the predicted amodal mask and the ground truth over the entire object region, and occ mIoU, which is computed exclusively over the occluded portion and directly reflects the model's ability to complete invisible regions. Experiments are conducted under two input settings. The \emph{Oracle} setting uses the ground-truth visible mask as input, measuring the model's upper-bound capacity. The \emph{Standard} setting uses visible masks predicted by the upstream AISFormer~\cite{aisformer2022} segmentation model, reflecting end-to-end performance under practical deployment conditions.

GRASP adopts a frozen DINOv2-ViT-B/14~\cite{oquab2024dinov2} as the visual encoder, concatenating patch tokens from the last four Transformer layers and projecting them into a 768-dimensional feature space. Input images are cropped to $224 \times 224$. SPM maintains 128 learnable shape prototypes with a single cross-attention layer. Spatial ARG contains only two learnable scalar parameters ($\alpha$ and $\beta$), generating per-position gate weights by applying a sigmoid transformation to the normalized SDF values. The model is trained with the AdamW optimizer~\cite{loshchilov2019adamw} using an initial learning rate of $1\times10^{-4}$ with cosine decay and a batch size of 32. KINS is trained for 11 epochs ($\sim$56K steps) and COCOA for 6 epochs ($\sim$8K steps). The entire framework comprises approximately 26M trainable parameters and 86M frozen parameters (DINOv2-ViT-B/14), and all experiments are conducted on a single NVIDIA RTX 5880 Ada GPU.

\subsection{Comparison with State-of-the-Art Methods}

\begin{table*}[!t]
\centering
\caption{Standard-setting comparison on KINS and COCOA. All methods use predicted visible masks as input. We report mIoU (\%) for full and occluded regions, along with total parameters and per-instance inference latency. Best results are \textbf{bold}.}
\label{tab:sota}
\renewcommand{\arraystretch}{1.15}
\begin{tabular}{l l c c c c c c}
\toprule
\textbf{Method} & \textbf{Prior Type} & \textbf{Params} & \textbf{Per-inst.} & \textbf{KINS full} & \textbf{KINS occ} & \textbf{COCOA full} & \textbf{COCOA occ} \\
\midrule
PCNet-M~\cite{zhan2020deocclusion}  & None     & --   & --          & 78.02 & 38.14 & 76.91 & 20.34 \\
AISFormer~\cite{aisformer2022}      & None     & --   & --          & 81.53 & 48.54 & 72.69 & 13.75 \\
VRSP~\cite{xiao2021amodal}          & Codebook & --   & --          & 80.70 & 47.33 & 78.98 & 22.92 \\
C2F-Seg~\cite{gao2023c2f}           & Codebook & 325M & 28.93\,ms   & 82.22$^\dagger$ & 53.60 & 80.28 & 27.71 \\
\textbf{GRASP (Ours)} & \textbf{Prototype} & \textbf{112M$^\ast$} & \textbf{9.78\,ms} & \textbf{82.37} & \textbf{55.24} & \textbf{84.92} & \textbf{39.47} \\
\bottomrule
\end{tabular}

\vspace{2pt}
{\footnotesize $^\ast$\,112M total, of which 26M trainable and 86M frozen (DINOv2-ViT-B/14). $^\dagger$\,C2F-Seg KINS full includes post-processing (pred\,$\cup$\,predicted VM); GRASP reports raw predictions without post-processing. Per-instance latency measured on a single NVIDIA RTX 5880 Ada GPU, batch size 1, averaged over 100 images ($\sim$1{,}300 instances); data I/O and post-processing excluded. Baseline accuracy numbers from respective original papers.}
\end{table*}

\begin{table}[!t]
\centering
\caption{Oracle upper-bound comparison on KINS and COCOA. Methods are grouped by input type to clarify protocol differences. Post-processing (pred\,$\cup$\,VM) is applied to GT~VM results. Best results per block are \textbf{bold}.}
\label{tab:oracle}
\renewcommand{\arraystretch}{1.15}
\resizebox{\columnwidth}{!}{%
\begin{tabular}{l l c c c c}
\toprule
\textbf{Method} & \textbf{Input} & \textbf{KINS full} & \textbf{KINS occ} & \textbf{COCOA full} & \textbf{COCOA occ} \\
\midrule
\multicolumn{6}{l}{\textit{Ground-truth visible mask:}} \\[2pt]
C2F-Seg~\cite{gao2023c2f}  & GT VM   & 87.89 & 57.60 & 87.13 & 36.55 \\
\textbf{GRASP}               & \textbf{GT VM} & \textbf{90.49} & \textbf{62.59} & \textbf{89.44} & \textbf{41.68} \\
\midrule
\multicolumn{6}{l}{\textit{Ground-truth bounding box:}} \\[2pt]
pix2gestalt~\cite{ozguroglu2024pix2gestalt} & GT bbox & 81.45 & -- & 79.08 & -- \\
SAMBA~\cite{liu2025samba}   & GT bbox & 88.47 & -- & 81.82$^\ddagger$ & -- \\
\bottomrule
\end{tabular}%
}

\vspace{2pt}
{\footnotesize $^\ddagger$\,COCOA-cls split, not directly comparable to the standard COCOA split used by other methods. SAMBA and pix2gestalt do not report occ mIoU.}
\end{table}

Table~\ref{tab:sota} presents the Standard-setting comparison, where all methods use predicted visible masks as input. GRASP achieves 82.37 and 55.24 in full and occ mIoU on KINS, surpassing the strongest codebook-based method~\cite{gao2023c2f} by +1.64 in occ mIoU. On COCOA, GRASP attains 84.92 and 39.47, outperforming the same method by 11.76 points in occ mIoU. The substantially larger gain on COCOA indicates that when object category diversity increases, the shape priors constructed by SPM through weighted prototype combination offer broader coverage than fixed-capacity codebook retrieval~\cite{gao2023c2f, esser2021taming}.

On KINS, the full mIoU margin is narrow (+0.15), but this comparison is asymmetric. The reported 82.22 of C2F-Seg~\cite{gao2023c2f} includes post-processing (pred\,$\cup$\,predicted VM), while GRASP's 82.37 is the raw prediction without any post-processing. The occ mIoU improvement of +1.64, combined with a 65\% parameter reduction and $3\times$ inference speedup, demonstrates that GRASP's advantage lies precisely in occlusion completion, the core challenge of amodal segmentation.

GRASP also offers clear advantages in computational efficiency. As shown in Table~\ref{tab:sota}, GRASP runs at 9.78\,ms per instance with 112M total parameters, approximately $3\times$ faster than the 28.93\,ms of C2F-Seg~\cite{gao2023c2f}. This gap arises because SPM completes shape prior construction through a single cross-attention layer, whereas C2F-Seg~\cite{gao2023c2f} requires multiple rounds of MaskGIT iterative decoding~\cite{esser2021taming}.

Table~\ref{tab:oracle} reports the Oracle upper-bound comparison, with methods grouped by input type for transparent protocol comparison. Under the ground-truth visible mask protocol, GRASP outperforms C2F-Seg~\cite{gao2023c2f} by +2.60/+4.99 on KINS and +2.31/+5.13 on COCOA in full/occ mIoU, confirming that the performance gap widens when upstream detector noise is removed. The gap between Oracle and Standard reflects the influence of upstream detector quality on final performance, which we analyze in detail in Section~IV-D2.

Recent foundation-model-based approaches such as SAMBA~\cite{liu2025samba} achieve strong full mIoU through billion-scale pretraining with bounding box input, representing a different design philosophy from lightweight, task-specific methods. As SAMBA uses a different input protocol (ground-truth bounding box rather than visible mask), the two setups differ in input granularity, making direct numerical comparison unreliable in either direction. Moreover, SAMBA reports COCOA-cls split results that differ from the standard COCOA split used by other methods, and does not report occ mIoU, precluding comparison on the metric most diagnostic of occlusion completion quality. Our work targets the complementary goal of achieving effective occlusion completion with minimal trainable parameters, and we focus detailed comparison on C2F-Seg~\cite{gao2023c2f} as the strongest open-source baseline within the prior-based paradigm.

\begin{figure*}[!t]
\centering
\includegraphics[width=\textwidth]{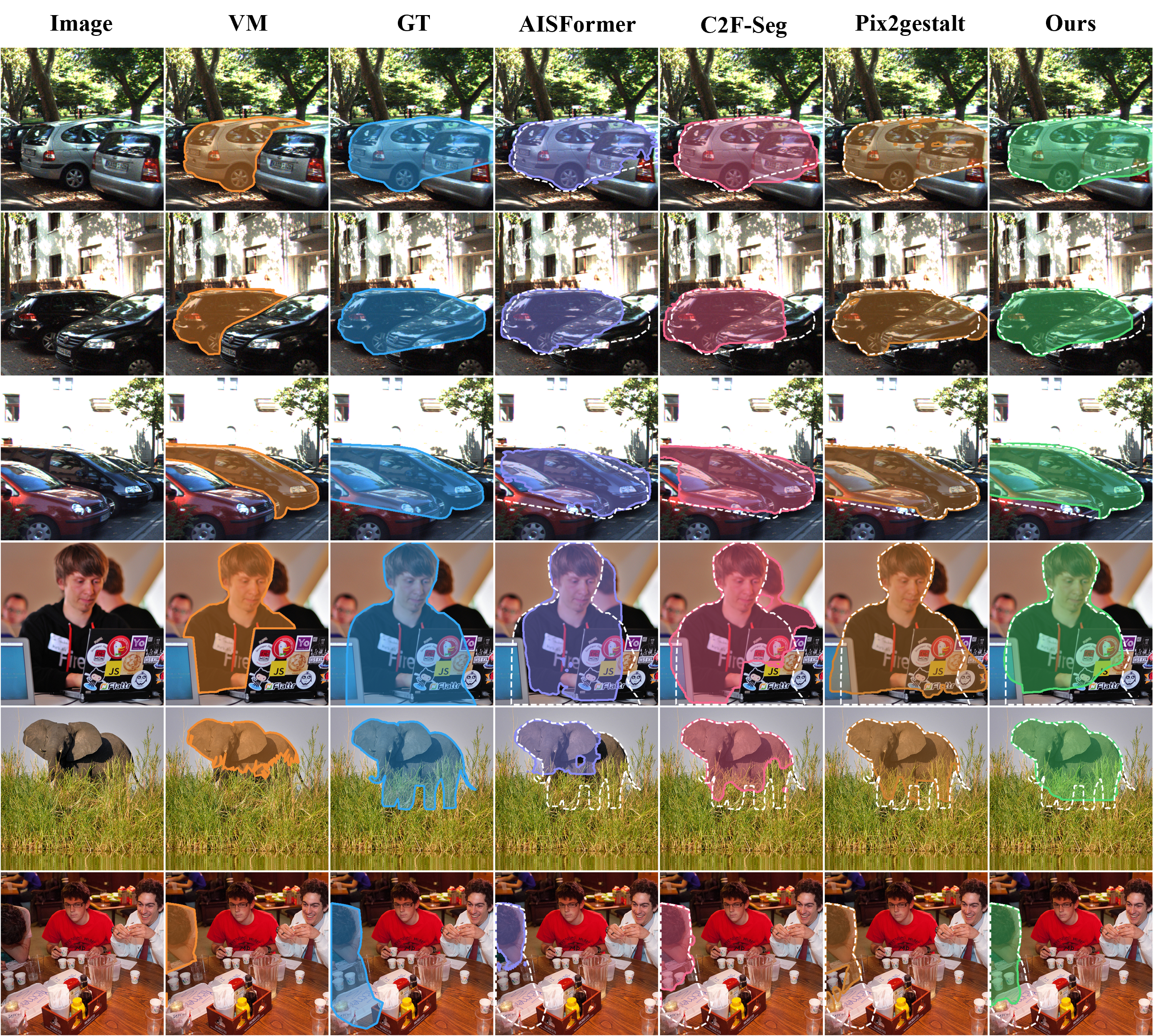}
\caption{Qualitative comparison with AISFormer~\cite{aisformer2022}, C2F-Seg~\cite{gao2023c2f}, and pix2gestalt~\cite{ozguroglu2024pix2gestalt} on KINS (rows 1--3) and COCOA (rows 4--6). Each row shows, from left to right, the input image, the visible mask in orange, the ground-truth amodal mask in blue, and the predicted amodal masks of AISFormer, C2F-Seg, pix2gestalt, and GRASP, with the predicted amodal contour overlaid as a dashed white curve.}
\label{fig:qualitative_compare}
\end{figure*}

As shown in Fig.~\ref{fig:qualitative_compare}, GRASP produces tighter amodal contours along occluded boundaries than AISFormer~\cite{aisformer2022}, C2F-Seg~\cite{gao2023c2f}, and pix2gestalt~\cite{ozguroglu2024pix2gestalt} across both vehicle and natural-scene categories. The baseline methods tend to over-extend predictions into background regions when the visible mask provides limited boundary evidence, whereas the contours produced by GRASP remain close to the true object silhouette even under heavy occlusion. This behavioral difference is consistent with the role of the spatial reliability gate, which suppresses shape prior injection inside well-supported visible regions and concentrates corrections within the genuinely occluded area, preventing the kind of unconstrained outward growth observed in the baseline predictions.

\subsection{Ablation Study}

\begin{table}[!t]
\centering
\caption{Module ablation on KINS Oracle. Inference-time gate intervention without retraining. PP applied.}
\label{tab:ablation}
\renewcommand{\arraystretch}{1.15}
\begin{tabular}{l c c c c}
\toprule
\textbf{Configuration} & \textbf{full} & \textbf{occ} & \textbf{$\Delta$full} & \textbf{$\Delta$occ} \\
\midrule
\textbf{Full (SPM + Spatial ARG)} & \textbf{90.49} & \textbf{62.59} & ref & ref \\
No SPM (gate=0)                   & 64.22 & 37.29 & $-$26.27 & $-$25.30 \\
SPM, no ARG (gate=0.5)            & 85.63 & 56.31 & $-$4.86  & $-$6.28  \\
Uniform gate ($g$$\approx$1)      & 87.69 & 50.84 & $-$2.80  & $-$11.75 \\
\bottomrule
\end{tabular}
\end{table}

To validate the effectiveness of each component in GRASP, we design four ablation configurations under the KINS Oracle setting. The experiments isolate individual modules by directly intervening on the gate values at inference time, without retraining the model. Table~\ref{tab:ablation} reports the full and occ mIoU for each configuration.

We first examine the contribution of SPM. Setting the gate to zero to completely block shape prior injection causes the occ mIoU to drop from 62.59 to 37.29, a decrease of 25.30 points, confirming that shape priors are the primary driver of occlusion completion.

We then examine the role of ARG adaptive modulation. When SPM is retained but the gate values at all positions are uniformly fixed to 0.5, the occ mIoU decreases to 56.31, a drop of 6.28 points. Although SPM still provides shape priors (56.31 is far above the 37.29 of No SPM), the lack of spatial guidance prevents the model from differentially allocating injection intensity according to occlusion depth.

The fourth row reveals a more critical finding. Removing spatial guidance and setting the gate to a near-uniform high value ($g$$\approx$1) causes the occ mIoU to plummet to 50.84 ($-$11.75), even 5.47 points below the fixed gate=0.5 configuration. This counter-intuitive result demonstrates that injecting shape priors at full intensity across all positions is actively harmful: accurate predictions in visible regions are disrupted by the prior information, while occluded regions receive no targeted enhancement.

Taken together, the core conclusion from Table~\ref{tab:ablation} is that the value of ARG lies not in controlling the magnitude of SPM injection, but in its \emph{spatial allocation}. Directing corrections precisely toward occluded regions while protecting visible regions from interference is the fundamental function of Spatial ARG. Since these ablations intervene on a jointly trained model, the observed drops include distribution mismatch in downstream layers and thus represent upper bounds on each module's isolated effect. In particular, setting gate=0 removes the shape prior signal that downstream decoder layers were trained to expect, amplifying the apparent contribution of SPM beyond its true isolated gain.

\subsection{Further Analysis}

\subsubsection{Stratified Occlusion Analysis}

\begin{table}[!t]
\centering
\caption{Stratified occlusion analysis on KINS Oracle. Occ mIoU reported by occlusion severity (occ\_area / amodal\_area). PP applied.}
\label{tab:stratified}
\renewcommand{\arraystretch}{1.15}
\begin{tabular}{l r c c c}
\toprule
\textbf{Occ Bin} & \textbf{$N$} & \textbf{Full Model} & \textbf{No SPM} & \textbf{$\Delta$SPM} \\
\midrule
0--25\%   & 14{,}142 & 55.21 & 20.10 & +35.11 \\
25--50\%  & 16{,}861 & 70.58 & 41.04 & +29.54 \\
50--75\%  & 14{,}281 & 64.58 & 48.29 & +16.29 \\
75--100\% &  4{,}959 & 50.71 & 41.84 &  +8.87 \\
\midrule
\textbf{Overall} & \textbf{50{,}243} & \textbf{62.59} & \textbf{37.29} & \textbf{+25.30} \\
\bottomrule
\end{tabular}
\end{table}

To further examine the contribution of SPM under varying degrees of occlusion, we divide the occluded instances in the KINS Oracle test set into four equal-width bins according to the ratio of occluded area to amodal area. Table~\ref{tab:stratified} presents the occ mIoU comparison between the full model and the model without SPM for each bin.

SPM yields significant improvements across all occlusion levels. The largest gain appears in the lightest occlusion bin (0--25\%), reaching +35.11 points, because sufficient visible context enables the visual tokens as queries to accurately combine matching shape priors from the prototype bank. As the occlusion ratio increases and available visual cues diminish, the gains decline monotonically to +29.54 (25--50\%), +16.29 (50--75\%), and +8.87 (75--100\%). Even under severe occlusion exceeding 75\%, SPM still contributes a +8.87-point improvement, indicating that the prototype bank provides effective shape constraints even when visible information is highly scarce. We focus the stratified analysis on KINS because its large test set ($N{=}50{,}243$ occluded instances) and homogeneous vehicle composition yield statistically reliable per-bin estimates. The COCOA test set, which contains roughly $3{,}800$ instances spread over $80$ categories, produces high per-bin variance once divided into four occlusion bins and makes bin-wise comparisons considerably less informative.

Aggregating the average Spatial ARG gate values across all 92{,}492 test instances reveals that the gate values increase monotonically with occlusion ratio: 0.610 for the 0--25\% bin, 0.634 for 25--50\%, 0.670 for 50--75\%, and 0.701 for 75--100\%. The standard deviation decreases concurrently (0.036$\to$0.016), indicating more consistent gating behavior at higher occlusion levels. On the $16\times16$ token grid, the average gate value is 0.448 at object centers, 0.733 at edges, and 0.793 at corners, exhibiting an outward-increasing spatial pattern that is fully consistent with the SDF-driven spatial allocation design.

\subsubsection{Upstream Detector Quality Analysis}

\begin{table}[!ht]
\centering
\caption{SPM gain versus upstream VM quality on KINS. Only occluded instances are included ($N$=32{,}360). VM IoU denotes the per-instance IoU between AISFormer-predicted VM and GT VM. Results are from single-pass inference; the gap relative to Table~\ref{tab:sota} (55.24) reflects the two-pass iterative refinement gain.}
\label{tab:vm_quality}
\renewcommand{\arraystretch}{1.15}
\begin{tabular}{l r c c c}
\toprule
\textbf{VM IoU} & \textbf{$N$} & \textbf{Occ SPM} & \textbf{Occ NoSPM} & \textbf{$\Delta$occ} \\
\midrule
50--65\%  & 9{,}725 & 46.25 & 39.29 & +6.96  \\
65--75\%  & 7{,}953 & 50.54 & 38.48 & +12.06 \\
75--85\%  & 9{,}117 & 58.62 & 37.85 & +20.77 \\
85--95\%  & 5{,}420 & 60.92 & 34.53 & +26.39 \\
95--100\% &    145  & 55.09 & 15.63 & +39.45 \\
\bottomrule
\end{tabular}
\end{table}

Table~\ref{tab:vm_quality} stratifies occluded instances by the prediction quality of the upstream detector. SPM gain increases monotonically with VM quality: $\Delta$occ grows from +6.96 in the 50--65\% bin to +26.39 in the 85--95\% bin, a $3.8\times$ amplification. In the 95--100\% bin ($N$=145), $\Delta$occ rises further to +39.45, though this data point should be interpreted with caution given the small sample size. In stark contrast, the No-SPM baseline shows no improvement and in fact degrades with better VM quality, with occ mIoU declining from 39.29 to 34.53 across the main bins. In the 95--100\% bin ($N$=145), No-SPM occ mIoU drops sharply to 15.63, though the very small sample size limits the reliability of this estimate. This demonstrates that the visual backbone alone cannot translate better visible masks into better occlusion completion; only the shape prior query mechanism provided by SPM establishes this link. This finding implies that as upstream detectors continue to improve, GRASP's performance will directly benefit.

An alternative interpretation of SPM's effect is that it merely performs boundary smoothing or local extrapolation from the visible mask edge, rather than drawing on category-level shape priors. This boundary-interpolation hypothesis predicts that SPM's contribution should be insensitive to upstream VM quality, since sharper edges should not yield disproportionately more occlusion completion. Table~\ref{tab:vm_quality} falsifies this prediction. SPM gain scales monotonically with VM quality (+6.96 $\to$ +26.39, a $3.8\times$ amplification), while the No-SPM baseline exhibits no such scaling and in fact slightly degrades as VM quality improves. Such a pattern is consistent only with a mechanism that uses the visible mask to \emph{query} stored shape priors, not one that \emph{interpolates} from its boundary.

\subsubsection{Cross-Protocol Performance Gap Analysis}

A noticeable asymmetry exists in GRASP's Standard-setting gains over C2F-Seg~\cite{gao2023c2f}: occ mIoU improves by +11.76 on COCOA but only +1.64 on KINS. This difference is primarily attributable to the cropping protocol rather than a limitation of the method itself. Under the Standard setting, KINS crops each instance using the bounding box of the \emph{visible} region, which systematically excludes deeply occluded areas that extend beyond the visible boundary. Since SPM still provides substantial gains in deeply occluded regions (Table~\ref{tab:stratified}: $\Delta$occ=+8.87 in the 75--100\% bin) and these regions are more likely to be excluded by visible-box cropping, the truncation disproportionately suppresses SPM's advantage on KINS. In contrast, COCOA uses the amodal bounding box for cropping, preserving the full occlusion context and allowing SPM's shape priors to be fully utilized.

The Oracle-to-Standard performance gap further corroborates this analysis. On KINS, the occ mIoU gap between Oracle and Standard is 7.35 points (62.59$\to$55.24), which is $3.3\times$ larger than the corresponding gap on COCOA (2.21 points, 41.68$\to$39.47). Diagnostic experiments indicate that approximately 69\% of the KINS Oracle-Standard gap is attributable to crop misalignment (the visible bounding box excludes occluded regions), while the remaining 31\% arises from VM prediction noise. This decomposition is obtained by separately substituting the ground-truth amodal bounding box and the ground-truth visible mask in place of their predicted counterparts at inference time, and attributing the recovered portion of the 7.35-point occ mIoU gap to each source. This decomposition suggests that adopting an amodal-aware cropping strategy (e.g., using a predicted amodal bounding box rather than the visible bounding box) could further release the potential of shape priors on datasets with visible-box cropping protocols. It is also worth noting that the post-processing step (pred\,$\cup$\,VM) used in the Oracle setting contributes substantially to full mIoU (e.g., +6.71 on COCOA Oracle), whereas under the Standard setting its effect is diminished because the predicted VM itself contains boundary noise.

The preceding analyses characterize when and where GRASP's gains arise. The following analysis shifts focus to why a minimal design of 128 learnable prototypes, a single cross-attention layer, and two scalar gate parameters suffices for effective shape prior construction, examining emergent behaviors that arise from training without explicit architectural or supervisory intervention.

\subsubsection{Implicit Geometric Encoding Analysis}

GRASP uses SDF geometry explicitly in Spatial ARG to control injection intensity (\emph{how much}), but does not explicitly inject geometric information into the prototype query within SPM (\emph{what}). A natural question arises: should SDF also be explicitly injected into the prototype query to enable tokens at different occlusion depths to route more precisely to different prototypes? We conduct two complementary experiments to address this question.

Adding an SDF query modulation before SPM cross-attention ($h = x + \text{sdf\_norm} \times d$, where $d$ is a 768-dimensional zero-initialized learnable direction vector adding only 768 parameters) results in a 1.52-point decrease in occ mIoU on COCOA Oracle (41.68$\to$40.16). The model not only fails to benefit from explicit geometric injection but slightly degrades.

\begin{table}[!ht]
\centering
\caption{Linear probe: SDF predictability from token features. R$^2$ and sign accuracy (SA) are reported for Ridge regression from token features to normalized SDF values. K = KINS, C = COCOA.}
\label{tab:linear_probe}
\renewcommand{\arraystretch}{1.15}
\begin{tabular}{l c c c c}
\toprule
\textbf{Feature Position} & \textbf{K-R$^2$} & \textbf{K-SA} & \textbf{C-R$^2$} & \textbf{C-SA} \\
\midrule
(A) DINOv2 proj (pre-CA) & 0.614 & 0.861 & 0.492 & 0.853 \\
\textbf{(B) VM Cross-Attn (post-CA)} & \textbf{0.774} & \textbf{0.904} & \textbf{0.629} & \textbf{0.893} \\
(C) Random baseline & $-$0.001 & 0.720 & $-$0.002 & 0.736 \\
\midrule
$\Delta$ (B$-$A) & +0.160 & +0.043 & +0.138 & +0.040 \\
\bottomrule
\end{tabular}

\vspace{2pt}
{\footnotesize 2{,}000 instances per dataset, 512K tokens total. 80/20 split. Ridge regression.}
\end{table}

\begin{figure}[!t]
\centering
\includegraphics[width= \linewidth]{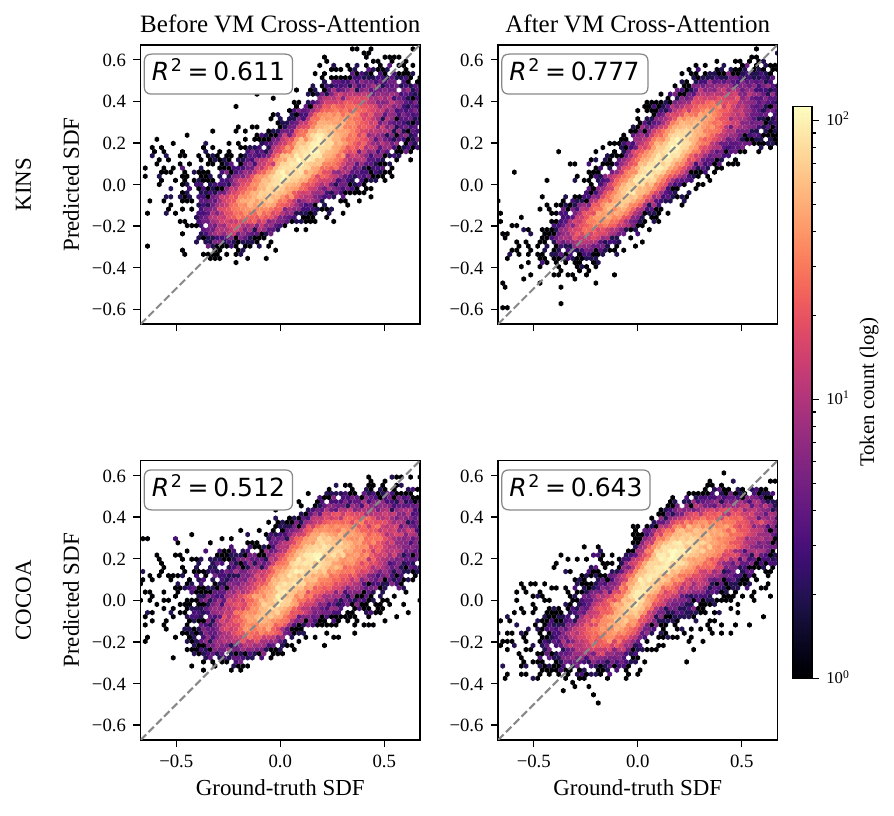}
\caption{Linear probing of SDF encoding in visual tokens. Hexbin density of ground-truth versus linearly probed SDF values, before (left) and after (right) the VM Cross-Attention layer, on KINS (top) and COCOA (bottom). The dashed line marks $y = x$. VM Cross-Attention substantially tightens the predicted distribution around the diagonal on both datasets, confirming that geometric occlusion structure is implicitly encoded in the visual tokens prior to SPM injection.}
\label{fig:linear_probe}
\end{figure}

To explain the negative result above, we employ linear probing to analyze whether visual tokens already implicitly encode SDF information. We freeze the trained model parameters, extract token features at two positions, (A)~before and (B)~after VM Cross-Attention, and train a Ridge regression model to predict the normalized SDF value at each token's spatial location. As shown in Table~\ref{tab:linear_probe} and Fig.~\ref{fig:linear_probe}, a linear regressor predicts token-level SDF from post-Cross-Attention features with R$^2$=0.774 on KINS and 0.629 on COCOA, substantially higher than from the raw DINOv2 features (0.614 and 0.492). Since geometric information is already encoded in the token representation with high fidelity, when these tokens serve as queries in SPM cross-attention, they naturally attend to different prototypes based on their inherent geometric context. This justifies omitting explicit SDF concatenation at the SPM input.

This finding can also be understood from a noise propagation perspective. In the gate pathway (Spatial ARG), the sigmoid function has a Lipschitz constant of 1/4, so VM prediction noise $\eta$ is compressed: $\|g_{\mathrm{noisy}} - g_{\mathrm{true}}\| \leq (\|\alpha\|/4) \cdot \|\eta_{\mathrm{norm}}\|$, and the output is always bounded in $[0, 1]$. This bounded compression ensures that even noisy visible masks produce spatially meaningful gate distributions, whereas injecting noisy SDF values directly into attention logits lacks such a built-in stability guarantee.

The implicit geometric encoding revealed above means that tokens entering SPM cross-attention already carry occlusion-depth information, and SPM does in fact exploit this signal. On COCOA, occluded tokens exhibit measurably more concentrated cross-attention over the prototype bank than visible tokens (Jensen--Shannon divergence of $0.293$ between the two groups, with Top-1 prototype weight $0.097$ versus $0.020$), indicating depth-dependent prototype utilization without any explicit routing supervision. This self-organized behavior is also robust to the prototype initialization scheme. Random Gaussian initialization yields essentially the same performance as K-Means initialization extracted from training-set shape features ($\Delta$occ\,$=$\,+0.20 on KINS Oracle), indicating that prototype specialization emerges from the amodal reconstruction loss alone rather than requiring careful initialization.

\subsubsection{Broader Comparison of Shape Prior Mechanisms}

A broader examination of existing shape prior mechanisms reveals a shared failure pattern under heavy occlusion. SAMBA~\cite{liu2025samba} introduces occlusion-rate-aware Mixture-of-Experts routing, explicitly motivated by the need for specialized processing at different occlusion levels. However, its own ablation (Table~4 in~\cite{liu2025samba}) shows that MoE provides no improvement at the heaviest evaluated level (FG-3: 70.8 with and without MoE + Distribution Loss), with all gains concentrated in lightly occluded instances. We additionally evaluated pix2gestalt~\cite{ozguroglu2024pix2gestalt} under its default GT-bbox protocol on 600 stratified KINS instances, obtaining an overall occ mIoU of 38.5\%, substantially below both GRASP (62.6\%) and C2F-Seg (57.6\%) under the GT~VM protocol. Combined with C2F-Seg's 4.99-point occ mIoU deficit relative to GRASP under Oracle evaluation (Table~\ref{tab:oracle}), these results suggest that existing approaches, whether discrete codebook lookup, diffusion-based reconstruction, or expert routing, share a common vulnerability when visible context becomes severely limited. GRASP's gated reliability-adaptive mechanism, combining soft prototype interpolation with spatially adaptive gating, offers an alternative design that sustains effectiveness in this critical regime.

The predominant reliance on full mIoU in amodal segmentation evaluation conflates visible-region accuracy with occlusion completion quality. Since the visible region typically constitutes 70--90\% of an object's amodal area, full mIoU is dominated by pixels that any competent segmentation model predicts correctly. This effect is evident in Table~\ref{tab:sota}: the full mIoU gap between AISFormer and C2F-Seg is only 0.69 points, whereas the corresponding occ mIoU gap reaches 5.06 points, suggesting that full mIoU substantially compresses differences in occlusion completion quality. We suggest that stratified occ mIoU reporting may serve as a more diagnostic complement to the standard full mIoU for this task.

\begin{figure}[!t]
\centering
\includegraphics[width=\linewidth]{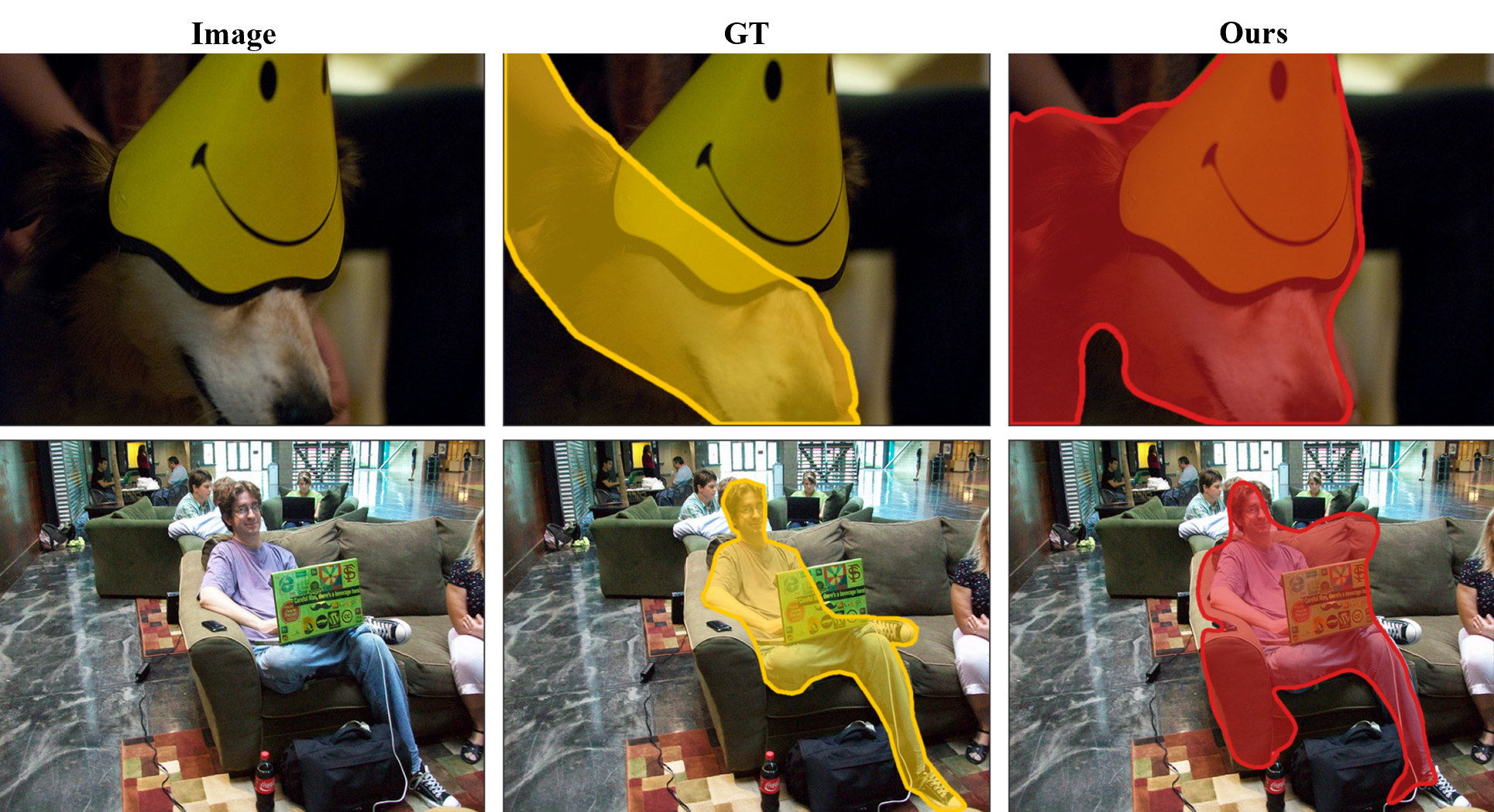}
\caption{Representative failure cases of GRASP on COCOA. Top row: semantic confusion, where the target arm is misidentified as the adjacent dog. Bottom row: over-extension, where fragmented visible regions cause the prediction to expand into surrounding furniture.}
\label{fig:qualitative}
\end{figure}

\subsection{Limitations and Failure Mode Analysis}
\label{sec:limitations}

While GRASP achieves consistent improvements across both benchmarks, an analysis of its failure cases reveals two underlying mechanisms that bound the framework's reliability, together with a separate, dataset-specific failure pattern on KINS that lies outside the scope of the proposed shape prior mechanism.

\textbf{Semantic confusion under visual dominance.} Fig.~\ref{fig:qualitative} (top row) illustrates a case where the target object is a human arm partially occluded by a dog wearing a smiling-face hat. The ground-truth amodal mask shows the arm's elongated contour extending from above, yet GRASP's prediction snaps entirely to the dog-and-hat composite shape and segments the wrong object altogether. This failure arises when the visible region of the target is small and spatially adjacent to a more visually dominant pattern. The prototype combination mechanism, which uses visible-region features as queries, attends to the salient dog-and-hat features rather than the subtle arm cues, causing the prediction to collapse onto a coherent but incorrect shape.

\textbf{Over-extension under fragmented evidence.} Fig.~\ref{fig:qualitative} (bottom row) shows a seated person whose visible region is partially occluded by handheld objects. GRASP's prediction dramatically over-extends beyond the person's true boundary, incorporating portions of the sofa and surrounding furniture into the predicted mask. The fragmented visible evidence fails to sufficiently constrain the shape completion process, and the combined shape prior, learned from typical human poses, expands to fill ambiguous regions without adequate boundary control.

Both failure modes stem from a common root cause, namely prior--likelihood imbalance. The Spatial Adaptive Reliability Gate decides \emph{how much} prior to inject at each position based solely on occlusion geometry, but has no mechanism to assess \emph{whether} the combined prior is semantically appropriate for the current instance. When visible evidence is weak, ambiguous, or dominated by adjacent objects, the gate continues to open in occluded regions and the combined prior dominates the prediction, pulling the output toward the nearest typical shape in the prototype bank. Addressing this limitation would require introducing an additional reliability signal that gauges the consistency between visible evidence and the combined prior, a direction we leave for future work.

\textbf{Failure modes specific to KINS.} A different pattern emerges on KINS, which we examined separately and chose not to include in Fig.~\ref{fig:qualitative}. As a subset of KITTI, KINS contains a high proportion of distant traffic participants whose instances span only a few dozen pixels in the original image and are further degraded by JPEG compression artifacts. For these instances, the dominant failure mode is boundary misalignment of small targets, where errors of a few pixels translate into substantial mIoU drops due to the small denominator. This failure mode does not stem from the shape prior mechanism but rather reflects the inherent difficulty of pixel-accurate segmentation on low-resolution inputs, and falls outside the scope of SPM or Spatial ARG.

\section{Conclusion}

In this paper, we have proposed GRASP, a framework that addresses the problem of occluded regions lacking pixel-level observations in amodal instance segmentation, from both the acquisition and injection of shape priors. For acquisition, learnable prototypes are combined via cross-attention into instance-level shape priors, replacing generative methods and fixed-capacity encoding spaces with a lightweight alternative. For injection, the signed distance field of the visible mask drives position-wise spatial gating, directing shape compensation toward occluded regions rather than distributing it uniformly across all positions. Extensive experiments on two benchmark datasets under multiple evaluation settings validate the effectiveness of the proposed method. Future work will explore extending the shape prior mechanism to video amodal segmentation. Another promising direction is to introduce reliability signals that gauge the consistency between visible evidence and the combined shape prior, which would mitigate the prior--likelihood imbalance observed in our failure analysis.


\end{document}